\def\year{2021}\relax
\documentclass[letterpaper]{article} 
\usepackage{aaai21}  
\usepackage{times}  
\usepackage{helvet} 
\usepackage{courier}  
\usepackage[hyphens]{url}  
\usepackage{bm}
\usepackage{graphicx} 
\usepackage{natbib}  
\usepackage{caption} 
\usepackage{subfigure}

\usepackage{amsmath}
\usepackage{booktabs}
\usepackage{algorithm}
\usepackage{amsfonts, amssymb}
\usepackage{algorithmic}
\usepackage{mathtools}
\usepackage{amsthm}

\title{Time Series Domain Adaptation via Sparse Associative Structure Alignment}

\author{
Ruichu Cai\textsuperscript{\rm 1,3},
Jiawei Chen\textsuperscript{\rm 1},
Zijian Li\textsuperscript{\rm 1},
Wei Chen\textsuperscript{\rm 1,3},
Keli Zhang\textsuperscript{\rm 2}, 
Junjian Ye\textsuperscript{\rm 2},
Zhuozhang Li\textsuperscript{\rm 1}, 
Xiaoyan Yang\textsuperscript{\rm 1}, 
Zhenjie Zhang\textsuperscript{\rm 1}\\
}
\affiliations{
\textsuperscript{\rm 1}Guangdong University of Technology\\
\textsuperscript{\rm 2}Huawei Noah's Ark Lab\\
\textsuperscript{\rm 3} Pazhou Lab\\
cairuichu@gdut.edu.cn, chenjiawei952@gmail.com, leizigin@gmail.com, chenweidelight@gmail.com, zhangkeli1@huawei.com, yejunjian@huawei.com, chokjohnlee@gmail.com, yangxiaoyan@gmail.com, zhenjie.zhang@pvoice.io

}

\begin{document}

\maketitle

\begin{abstract}
Domain adaptation on time series data is an important but challenging task. Most of the existing works in this area are based on the learning of the domain-invariant representation of the data with the help of restrictions like MMD. However, such extraction of the domain-invariant representation is a non-trivial task for time series data, due to the complex dependence among the timestamps. In detail, in the fully dependent time series, a small change of the time lags or the offsets may lead to difficulty in the domain invariant extraction. Fortunately, the stability of the causality inspired us to explore the domain invariant structure of the data. To reduce the difficulty in the discovery of causal structure, we relax it to the sparse associative structure and propose a novel sparse associative structure alignment model for domain adaptation. First, we generate the segment set to exclude the obstacle of offsets. Second, the intra-variables and inter-variables sparse attention mechanisms are devised to extract associative structure time-series data with considering time lags. Finally, the associative structure alignment is used to guide the transfer of knowledge from the source domain to the target one. Experimental studies not only verify the good performance of our methods on three real-world datasets but also provide some insightful discoveries on the transferred knowledge.

\end{abstract}

\section{Introduction}
Domain adaptation \cite{pan2009survey,long2015learning,cai2019learning}, utilizing both the labeled source domain data and the unlabeled target domain data, has a wide range of applications \cite{ganin2015unsupervised,lin2018neural}. To address the well-known phenomenon named \textit{domain shift}, a large number of methods have been proposed by exploring various assumptions between the source and target domains \cite{tzeng2014deep,cai2019learning,MDD_ICML_19}.  

One of the most widely used assumptions in domain adaptation is the existence of domain-invariant representation in both source and target domains. Since such an assumption has achieved great performance in non-time series data \cite{cai2019learning,wang2019transferable}, researchers have extended it to the time series data, by employing models like Recurrent Neural Network(RNN) \cite{mikolov2010recurrent} and variational RNN \cite{chung2015recurrent}, to learn representation from time series and using the gradient reversal layer (GRL) to align the representations learned from source and target time series data.

\begin{figure}
	\centering
	\includegraphics[width=\columnwidth]{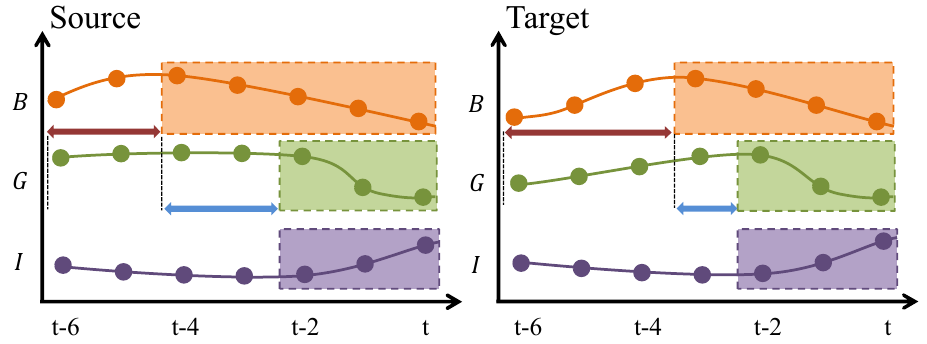}
	\caption{The illustration of the physiological mechanism in the human body among ``Blood glucose (B) $\downarrow$'', ``Glucagon (G) $\downarrow$'' and ``Insulin (I) $\uparrow$''. The decrease of ``Blood glucose'' leads to the decrease of ``Glucagon'' and the increase of ``Insulin''. The colored blocks denote the segments of change of variables. The different lengths of red double-head arrows denote different offsets. And the different lengths of blue double-head arrows denote different response times between ``Blood glucose'' and ``Glucagon''. Different response time means different time lags. (\textit{Best view in color.})}
	\label{fig:motivation1}
\end{figure}

However, extracting domain-invariant information from time series data is a challenging task. Existing methods \cite{da2020remaining,DBLP:conf/iclr/PurushothamCNL17}, which simply employ the RNN based feature extractor, essentially assume that the conditional distributions are equal \cite{pan2010domain}, i.e., $P_S(y|\varphi(x_1, x_2, \cdots, x_t)=P_T(y|\varphi(x_1, x_2, \cdots, x_t)$, in which $\varphi(\cdot)$ is the feature transformation mapping. This assumption works well in the static data but is difficult to satisfy in the time series data. Take Figure \ref{fig:motivation1} as an example, due to the complex dependency structure among the timestamps, even small discrepancies from different domains (e.g. offsets and varying of time lags) may result in the difficulty to learn the domain invariant representation. Furthermore, as for the multivariable time series data, variables are not always i.i.d. Existing methods for time series domain adaptation, which ignore the associative structure among variables, might suffer from overfitting.

\begin{figure}[ht]
	\centering
	\includegraphics[width=0.9\columnwidth,height=0.7\columnwidth]{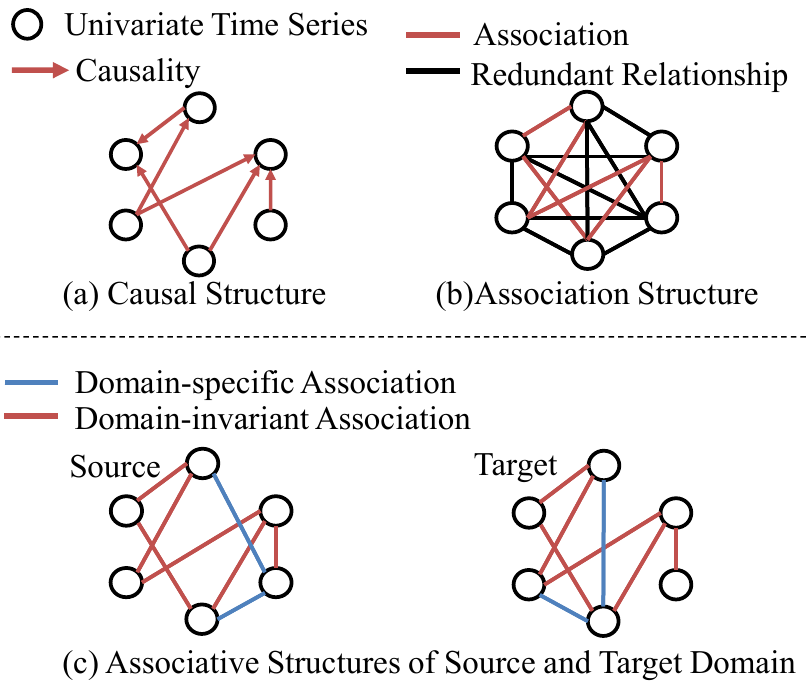}
	\caption{The illustration of various structures among six time series. (a) The causal structure of variables. (b) The existing methods take the conditional independence relationships into account and lead to overfitting. (c) Inspired by the stability of the causal mechanism, our method considers the stable and sparse associative structure among variables.}
	\label{fig:motivation2}
\end{figure}

Motivated by the toy example of Figure \ref{fig:motivation2}(a), data from the source and the target domains share the same stable causal structure (e.g., the physiological mechanism among ``Blood glucose$\downarrow$''(B), ``Glucagon$\downarrow$''(G) and ``Insulin$\uparrow$'' (I) shown in Figure \ref{fig:motivation1}), which is domain-invariant. However, as shown in Figure \ref{fig:motivation2}(b), the existing methods consider not only the ground truth associative structure but also the redundant relationships, which leads to overfitting. Since the causal structure from different domains is the same, time series data from the source and the target domains also share a similar associative structure. Figure \ref{fig:motivation2}(c) gives another insightful example, showing that considering domain-invariant associative structure and excluding domain-specific associations is important and can make the model robust and generalizable.

However, how to construct the associative structure among variables in time series data is another challenge, which is caused by the well-known discrepancy like time lags and offsets.
According to the physiological mechanism of the human body, the decrease of ``Blood glucose'' leads to the decrease of ``Glucagon'' and the increase of ``Insulin'', and the response time of the physiological mechanism varies with ages and races, resulting in different time lags (i.e., different length of blue double-head arrows from the source and the target domains in Figure \ref{fig:motivation1}). Giving another example, let source domain data and target domain data be sampled from the elder and the younger patients respectively, the response time of the elder patients is longer than the younger ones. 
At the same time, the same mechanism often happens with varying start points as indicated by different offsets from a different domain (i.e., the different length of red double-head arrows in Figure \ref{fig:motivation1}). Existing work, which simply adopting RNNs as feature extractors to extract the domain-invariant representation, can not exclude the negative influence of time lags and offsets and further fails to extract the associative structure.

Based on the above intuition, we propose the sparse associative structure alignment (\textbf{SASA}) approach for time series domain adaptation. The main challenges of SASA can be summarized into two folds. (1) How to get rid of the obstruction of time lags and offsets to extract the sparse associative structure? (2) How to extract the common associative structure and further extract the domain-invariant representation? 
To address these problems, first, we propose the adaptive segment summarization to ease the obstacle of offsets. Second, the proposed model extracts the sparse associative structure of the time series data via intra-variables and the inter-variables attention mechanism. 
Third, our model transfers the sparse associative structure from the source domain to the target domain by simply aligning the structure. Extensive experimental studies demonstrate that our \textbf{SASA} model outperforms the state-of-the-art time series unsupervised domain adaptation methods on three real world datasets.

\section{Related Works}
In this section, we mainly focus on the existing techniques on unsupervised domain adaptation as well as time series domain adaptation.

\begin{figure*}
		\centering
		\includegraphics[width=2\columnwidth]{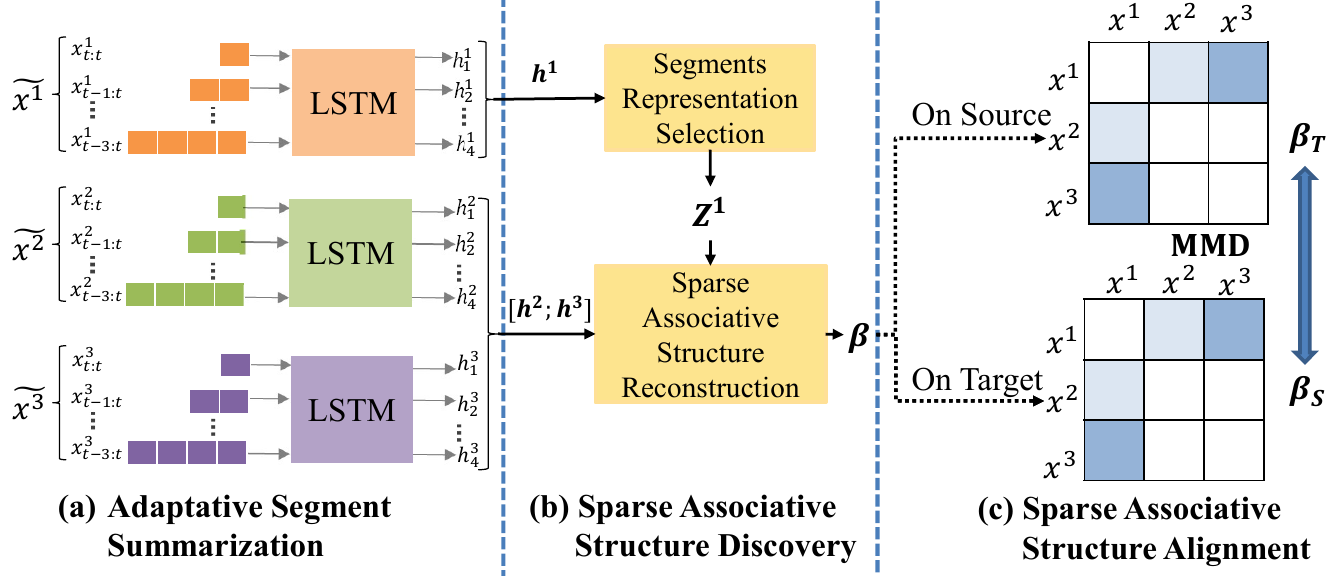}
		\caption{The framework of the sparse associative structure alignment model. (a) Adaptative segment summarization process with variable-specific LSTM. (b) Sparse associative structure discovery via intra-variables and inter-variables attention mechanism. (c) Sparse associative structure alignment between the source and the target domain. (\textit{Best view in color.})}
		\label{fig:model_1}
\end{figure*}

\noindent\textbf{Domain Adaptation on Non-Time Series Data. }
	The mainstream methods of unsupervised domain adaptation \cite{pan2010domain,wei2016deep,wei2018feature,wen2019exploiting} aim to extract the domain invariant representation between the source and the target domains. Maximum Mean Discrepancy (\textbf{MMD}) is one of the most popular methods by using kernel-reproducing Hilbert space \cite{tzeng2014deep}. Sun et al. \cite{sun2016return} propose second-order statistics for unsupervised domain adaptation. And Long et al. \cite{long2015learning} reduce the domain discrepancy by using an optimal multi-kernel selection method.
	
	Another essential approach to unsupervised domain adaptation is to extract the domain-invariant representation by borrowing the idea of generative adversarial networks \cite{goodfellow2014generative}. Ganin et al. \cite{ganin2015unsupervised} introduce a gradient reversal layer to fool the domain classifier and further extract the domain-invariant representation. Tzeng et al. \cite{tzeng2017adversarial} propose a novel unified framework for adversarial domain adaptation. Recently, considering the fine-grained alignment and aiming to prevent the false alignment. Xie et al. \cite{xie2018learning} address the unsupervised domain adaptation problem by aligning the centroid for each class in source and target domains with the help of pseudo labels.
	
	Under the causality view over the variables, the domain adaptation scenario can be determined by the causal mechanism. Three scenarios including target shift, condition shift, and generalized target shift, are discussed by Zhang et al. \cite{zhang2013domain}. Based on the former, Germain et al.\cite{germain2016new} and Zhang et al. \cite{zhang2015multi} investigate more on the generalized target shift in the context of domain adaptation. Recently, following the causal model of the data generation process, Cai et al. \cite{cai2019learning} address this problem by extracting the disentangled semantic representation on the recovered latent space.
	
	In this paper, we study the problem of unsupervised domain adaptation for time series data. Our \textbf{SASA} method is first inspired by the causal mechanism from observed data. And we further consider a more relaxed sparse associative structure since any two variables contain causal structure also have associative structure.
	
	\noindent\textbf{Domain Adaptation on Time Series data. }
	Recently, unsupervised domain adaptation on time series data has received more and more attention. 
	Da Costa et al. \cite{da2020remaining} employ the most straightforward method and simply replace with feature extractor with RNN based feature extractors to extract the representation of time series data. Purushotham et al. \cite{DBLP:conf/iclr/PurushothamCNL17} use variational RNN \cite{chung2015recurrent} to extract the latent representations of time series. There are limited works on time series domain adaptation. 
	One possible solution is the direct extension of the unsupervised domain adaptation methods on non-time series data to the time series data. However, this straightforward method might not work in time series data, since it's difficult to align the conditional distribution of the observed data in all timestamps. 
	
	Since the existing methods \cite{da2020remaining,DBLP:conf/iclr/PurushothamCNL17} for time series domain adaptation cannot well align the condition distribution of the time series data, we propose a novel domain adaptation method for time series data, which aims to distill the sparse associative structure and filter the domain-specific information.

\section{Sparse Associative Structure Alignment}

	In this section, we elaborate our Sparse Associative Structure Alignment (\textbf{SASA}) Model that distills the sparse associative structure and extracts the domain-invariant information from time series data. In this section,  we first formulate the problem. Then, we provide the details of our model. 

\subsection{Problem Formulation and Overview}
    In this work, we focus on the problem of unsupervised domain adaptation for time series data. We let $\bm{x}=\{\bm{x}_{t-N+1},\cdots, \bm{x}_{t-1},  \bm{x}_t\}$ denote a multivariate time series sample with $N$ time steps, where $\bm{x}_t\in\mathbb{R}^M$, and $y\in\mathbb{R}$ is the corresponding label. 
    We assume that $P_S(\bm{x},y)$ and $P_T(\bm{x},y)$ are different distributions from the source and the target domains but are generated from a shared causal mechanism. 
    Since the two variable sets generated by a same causal structure should share the same associative structure, $P_S(\bm{x},y)$ and $P_T(\bm{x},y)$ share the same associative structure. 
    $(\mathcal{X}_S, \mathcal{Y}_S)$ and $(\mathcal{X}_T, \mathcal{Y}_T)$, which are sampled from $P_S(\bm{x},y)$ and $P_T(\bm{x},y)$ respectively, denote the source and target domain dataset. 
    Then we further assume that each source domain sample $\bm{x}_S$ comes with $y_S$, while the target domain has no labeled sample. Our goal is to devise a predictive model that can predict $y_T$ given time series sample $\bm{x}_T$ from the target domain.

    To achieve this goal, we aim to extract the domain-invariant representation in the form of associative structure. This solution is inspired by intuition that the causal mechanism is invariant across the domains. Due to the complexity of learning causal structure, we relax the causal structure to the sparse associative structure. Considering that the offsets vary with different domains and hinder the model from extracting the domain-invariant associative structure, we first elaborate on how to obtain the fine-grain segments of time series data to ease the obstacle of the offsets. Sequentially, we reconstruct the associative structures via the intra-variables attention mechanism and the inter-variables attention mechanisms with considering time lags from different domains. Different from the existing works that align the feature from different domains, our \textbf{SASA} model aligns the common associative structures from different domains to indirectly extract the domain-invariant representation.

    \subsection{Adaptive Segment Summarization}
    In this subsection, we will elaborate on how to obtain the candidate segments to remove the obstacle of offsets.
    As shown in Figure \ref{fig:motivation1}, the orange blocks, whose duration varies with different domains, denote the segment of the change of variable `B'. 
    Existing methods, which take the whole time series data as input, can not accurately capture when a segment starts and when a variable affects the others, i.e., the sphere of influence of any variables. Therefore, these methods can not address the noise of offsets (i.e., the duration between the start point of time series and the start point of a segment).
    
    To address the aforementioned problem, we first propose the adaptive segment summarization. To obtain the candidate segments of $i$-th time series $\bm{x^i}=\{\bm{x}^i_{t-N+1}, \cdots, \bm{x}^i_{t-1},\bm{x}^i_t\}$, we construct multiple segments with different length for each variable $\bm{x}^i$. We have:
    \begin{small}
	\begin{equation}
    	\widetilde{\bm{x}^i}=\{\bm{x}^i_{t:t},\bm{x}^i_{t-1:t},\cdots,\bm{x}^i_{t-{\tau}+1:t},\cdots,\bm{x}^i_{t-N+1:t} \}.
	\end{equation}
	\end{small}
	Motivated by RIM \cite{goyal2019recurrent}, we allocate an independent LSTM for each variable. In detail, given a segment of $i$-th variable with $\tau$ time steps, we have:
	\begin{small}
	\begin{equation}
	     h_{\tau}^i=f(\bm{x}^i_{t-\tau+1:t};\bm{\theta}^i),
	\end{equation}
	\end{small}
	in which $\bm{\theta^i}$ denote the parameters of $i$-th LSTM. Note that the segments of any univariate time series $\bm{x^i}$ share the same LSTM, and finally we can obtain the segments representation set shown as follow:
	\begin{small}
	\begin{equation}
	    \bm{h}^i=\{{h}^i_{1},{h}^i_{2},\cdots,{h}^i_{{\tau}},\cdots,{h}^i_{N} \}.
	\end{equation}
	\end{small}
	 Since it's almost impossible to manually extract all the exact segments from the multivariable time series data, we first obtain the representation of all candidate segments via the aforementioned processing. The most suitable segment representations are selected and used to reconstruct the associative structure.
	
    \subsection{Sparse Associative Structure Discovery}
    In this section, we will introduce how to generalize the most exact segment representations and how to reconstruct the associative structure with the help of intra-variable attention mechanism and inter-variable attention mechanism respectively.

\subsubsection{Segments Representation Selection via Intra-Variables Attention Mechanism.}
In order to get rid of the obstacle brought from the offsets, we need to pay more attention to the exact segment representation among all the candidate segment representations with the help of the self-attention mechanism \cite{vaswani2017attention}. Formally, we calculate the weights of each segment of $\bm{x^i}$ as follow:
\begin{small}
\begin{equation}
	\begin{split}\label{equ:alpha}
	u^i_{\tau}&=\frac{1}{N}\sum_{k=1}^{N}\frac{(h^i_{\tau}\bm{W}^{Q})(h^i_k\bm{W}^{K})^\mathsf{T}}{\sqrt{d_h}},\\
	\bm{\alpha^i} &= \{\alpha_1^i, \alpha_2^i,\cdots,\alpha_{\tau}^i,\cdots,\alpha_N^i\}\\&=\text{sparsemax}(u^i_1,u^i_2,\cdots,u^i_{\tau},\cdots,u^i_N),
	\end{split}
\end{equation}
\end{small}
in which $\bm{W}^{Q}, \bm{W}^{K}$ are trainable projection parameters and $\sqrt{d_h}$ is the scaling factor. In order to obtain the sparse weights that represent specific segment representation clearly, we choose sparsemax \cite{martins2016softmax} to calculate the weights. The sparsemax is defined as $\text{sparsemax}(\bm{z})=\mathop{\arg \min_{\bm{p} \in  {\Delta}^{K-1}}}{||\bm{p}-\bm{z}||}^2$, which returns the Euclidean projection of vector $\bm{z}\in \mathbb{R}^K$ onto probability simplex ${\Delta}^{K-1}$.
As a result, we obtain the weighted segment representation of variable $\bm{x^i}$ as follow:
\begin{small}\begin{equation}
	\begin{split}
	Z^i = \sum_{{\tau}=1}^{N}\alpha_{\tau}^i\cdot (h_{\tau}^i\bm{W}^{V}),
	\end{split}
\end{equation}\end{small}
in which $\bm{W}^{V}$ is trainable projection parameter. Note that $\bm{\alpha}$ also denotes the probability of the length of a segment. For generalization, we also consider the case that the duration of a segment of a given variable varies with different domains. 
In this case, in order to reconstruct the associative structure more precisely, we minimize the maximum mean discrepancy (\textbf{MMD}) between $\bm{\alpha}$ from the source and the target domain to remove the obstacle of offsets. It restricts the duration of the segment from different domains to be similar, which contributes to extracting structure for transfer. 
Formally, we have:
\begin{small}
\begin{equation}
    \begin{split}
        \mathcal{L_{\alpha}}=\sum_{m=1}^M||\frac{1}{|\mathcal{X}_S|}\sum_{\bm{x}_S \in \mathcal{X}_S}  \bm{\alpha}_S^m - \frac{1}{|\mathcal{X}_T|}\sum_{\bm{x_T} \in \mathcal{X}_T} \bm{\alpha}_T^m||,
    \end{split}
\end{equation}
\end{small}
in which $\bm{\alpha}_S^m$ and $\bm{\alpha}_T^m$ denote the weights of segments of the $m$-th variable from the source and the target domains calculated by Equation (\ref{equ:alpha}).

\subsubsection{Sparse Associative Structure Reconstruction via Inter-variables Attention Mechanism.} 
With the help of the intra-variables attention mechanism, we extract the weighted segment representations despite the obstacle of offsets. Then we utilize these weighted segment representations to reconstruct the sparse associative structure among variables. 
So we propose the inter-variables attention mechanism to mine the associative structure among variables.

In this part, our goal is to reconstruct the associative structure among variables. Instead of using the self-attention mechanism in the intra-variables attention mechanism, we employ the ``referenced'' attention mechanism \cite{DBLP:journals/corr/BahdanauCB14}. One of the most straightforward methods to calculate the degree of correlation of variable $i$ and variable $j$ is shown as follow:
\begin{small}\begin{equation}\label{equ:no_time_lag}
    \bm{e}^{ij} = \frac{Z^i \cdot Z^j}{||Z^i|| \cdot ||Z^j||}
\end{equation}\end{small}
However, the associative structure calculated by Equation (\ref{equ:no_time_lag}) ignores the time lags from different domains between $i$ and $j$. Since Equation (\ref{equ:no_time_lag}) does not refer to time lags among variables, the associative structure might be falsely estimated.
In order to take the time lags into account, we calculate the degrees of association between variable $i$ and variable $j$ by:
\begin{small}\begin{equation}\label{equ:cor_no_norm}
    \begin{split}
        e^{ij}_\tau&=\frac{Z^i \cdot {h}^j_\tau}{||Z^i||\cdot ||h^j_\tau||} ,\\
        \bm{e}^{ij}&=\{e^{ij}_{1},e^{ij}_{2},\cdots,e^{ij}_{{\tau}},\cdots,e^{ij}_{N} \}.
    \end{split}
\end{equation}\end{small}
Then we normalize these degrees of association with sparsemax \cite{martins2016softmax}. Formally, we have:
\begin{small}\begin{equation}\label{equ:cor_norm}
    \begin{split}
        \bm{\beta}^{i} &=\{\bm{\beta}^{i1},\bm{\beta}^{i2},\cdots,\bm{\beta}^{ij},\cdots,\bm{\beta}^{iM}\}\\
        &= \text{sparsemax}(\{\bm{e}^{i1},\bm{e}^{i2},\cdots,\bm{e}^{ij},\cdots,\bm{e}^{iM}\})(j\neq i).
    \end{split}
\end{equation}\end{small}
Note that $\beta^{ij}_{\tau} \in \bm{\beta}^{i}$ denotes the associative strength between variable $i$ and variable $j$ with regard to segment duration of $\tau$. 

\subsection{Sparse Associative Structure Alignment}
In this subsection, we aim to extract the domain-invariant information for time series data with the help of the extracted associative structure from the source and the target domains.

We reconstruct the associative structure by Equation (\ref{equ:cor_no_norm}) and (\ref{equ:cor_norm}) taking the time lags into account. In order to extract the domain-invariant associative structure, we need to restrict the distance of the structure between the source and the target domains. Since $\bm{\beta}^{ij}$ can be seen as the associative strength distribution between $i$ and $j$, we turn the problem of structure distance measure to the distribution distance measure. In this paper, we borrow the idea of domain confuse network \cite{tzeng2014deep} and employ maximum mean discrepancy (\textbf{MMD}) for associative structure alignment. Formally, we have:
\begin{small}\begin{equation}
    \begin{split}
        \mathcal{L_{\beta}}=\sum^M_{m=1}||\frac{1}{|\mathcal{X}_S|}\sum_{\bm{x}_S \in \mathcal{X}_S}  \bm{\beta}_S^{m} - \frac{1}{|\mathcal{X}_T|}\sum_{\bm{x}_T \in \mathcal{X}_T} \bm{\beta}_T^{m}||.
    \end{split}
\end{equation}\end{small}
Note that we minimize the associative structure adjacent matrix instead of aligning the features like what \cite{tzeng2014deep} does.

\subsection{Model Summary}
\subsubsection{Task based Label Predictor.} We aim to obtain the domain-invariant representations which are combined with the sparse associative structure $\bm{\beta}$. In detail, we first calculate the associative structure representations of variable $j$, which is shown as follow:
\begin{small}\begin{equation}
    \begin{split}
        &\bm{U}^{ij} = \sum_{\tau=1}^{N}{\beta}^{ij}_\tau \cdot {h}^j_\tau ,\\
        &{U}^i = \sum_{m=1, m \neq i}^{M}\bm{U}^{im}.
    \end{split}
\end{equation}\end{small}
	
As a result, we can obtain the final representations by concatenating weighted segment representations and associative structure representations as follow:
	\begin{equation}
	\begin{split}
	H^i=\left[Z^i;U^i\right].
	\end{split}
	\end{equation}
For convenience, we describe the above process as:
\begin{small}\begin{equation}
	\begin{split}
	\bm{H} &= G_H(f(\bm{x};\bm{\Theta});\bm{W}^Q, \bm{W}^K, \bm{W}^V),\\
	\end{split}
\end{equation}\end{small}
in which $G_H$ actually denotes the feature extractor containing the aforementioned two kinds of attention mechanisms. $\bm{H}=[H^1;H^2;\cdots;H^M]$ denotes the final representation, we further let $\bm{\Theta}$ be the parameters of variable-specific LSTM.

After obtaining the final representation, we take $\bm{H}$ as the input of label classifier $G_y(\cdot;\bm{\phi})$ whose loss function is $\mathcal{L}_y$. For the classification problems, we employ cross-entropy as the label loss. For the regression problems, we employ RMSE as the label loss.
	
The label classifier with the trained optimal parameters is adapted to the target domains.
	\begin{small}\begin{equation}
	\begin{split}
	y_{pre} = G_y(G_H\left(f\left(\bm{x};\Theta\right);\bm{W}^Q, \bm{W}^K, \bm{W}^V\right),\bm{\phi}). 
	\end{split}
	\end{equation}\end{small}

\subsubsection{Objective Function.} 
The total loss of the proposed structure alignment model for time series domain adaptation is formulated as:
	\begin{small}\begin{equation}
	\begin{split}
	\mathcal{L}\left(\bm{\Theta}, \bm{W}^Q, \bm{W}^K,\bm{W}^V, \bm{\phi}\right) = \mathcal{L}_y + \omega (\mathcal{L}_{\alpha} + \mathcal{L}_\beta),
	\end{split}
	\end{equation}\end{small}
in which $\omega$ is hyper-parameter. 
	
Under the above objective function, our model is trained on the source and target domain using the following procedure:
	\begin{small}\begin{equation}
	\begin{split}
	&\left(\bm{\Theta}, \bm{W}^Q, \bm{W}^K, \bm{W}^V, \bm{\phi}\right)=\\
	&\mathop{\arg \min_{\bm{\Theta}, \bm{W}^Q, \bm{W}^K, \bm{W}^V, \bm{\phi}}}\mathcal{L}\left(\bm{\Theta}, \bm{W}^Q, \bm{W}^K, \bm{W}^V, \bm{\phi}\right).
	\end{split}
	\end{equation}\end{small}

\section{Experiments and Result}
	\subsection{Setup}
	
	\begin{table*}
	\small
		\centering
		\resizebox{\textwidth}{17mm}{
		    \scalebox{0.9}{
			\begin{tabular}{l|lllllllllllll}
				\hline
				Method&B$\rightarrow$T&G$\rightarrow$T&S$\rightarrow$T&T$\rightarrow$B&G$\rightarrow$B&S$\rightarrow$B&B$\rightarrow$G&T$\rightarrow$G&S$\rightarrow$G&B$\rightarrow$S&T$\rightarrow$S&G$\rightarrow$S&Avg \\
				\hline
				LSTM\_S2T  	           & 40.20 & 41.67 & 48.91 & 52.81 & 56.44 & 68.14 & 19.00 & 19.76 & 17.56 & 13.82 & 13.82 & 13.86 & 33.83 \\ 
				RDC            		   & 39.72 & 40.80 & 47.75 & 51.98 & 55.83 & 67.67 & 18.18 & 19.10 & 15.43 & 13.70 & 13.75 & 13.76 & 33.14 \\
				R-DANN                 & 39.93 & 40.98 & 46.16 & 52.72 & 55.65 & 66.47 & 18.00 & 18.47 & 15.18 & 13.82 & 13.78 & 13.79 & 32.91 \\
				VRADA                  & 38.12 & 38.69 & 45.29 & 52.14 & 54.51 & 64.41 & 17.30 & 17.95 & 14.63 & 13.80 & 13.90 & 13.80 & 32.04 \\
				\hline
				SASA-$\bm{\alpha}$             & 36.60 & 34.42 & 41.31 & 48.34 & 54.20 & 59.09 & 16.42 & 16.48 & 14.30 & 13.68 & 13.53 & 13.47 & 30.15\\
				SASA-$\bm{\beta}$             & 35.54 & 35.10 & 42.16 & 48.40 & 54.42 & 60.45 & 16.66 & 16.58 & 14.62 & 13.62 & 13.49 & 13.68 & 30.39\\
				
				SASA             & \textbf{34.26} & \textbf{33.84} & \textbf{40.91} & \textbf{48.15} & \textbf{54.14} & \textbf{56.80} & \textbf{16.40} & \textbf{15.41} & \textbf{14.23} & \textbf{13.49} & \textbf{13.46} & \textbf{13.38} & \textbf{29.54} \\
				
				\hline
				
		\end{tabular}}}
		
		\caption{RMSE on air quality prediction.}
		\label{tab:air_quality}
	\end{table*}
	
	\begin{table*}
		\centering
		\small
		\resizebox{\textwidth}{17mm}{
			\begin{tabular}{l|lllllllllllll}
				\hline
				Method&2$\rightarrow$1&3$\rightarrow$1&4$\rightarrow$1&1$\rightarrow$2&3$\rightarrow$2&4$\rightarrow$2&1$\rightarrow$3&2$\rightarrow$3&4$\rightarrow$3&1$\rightarrow$4&2$\rightarrow$4&3$\rightarrow$4&Avg \\
				\hline
				LSTM\_S2T 	           & 80.52 & 78.79 & 76.85 & 80.24 & 81.43 & 77.24 & 75.77 & 79.30 & 75.56 & 65.79 & 68.93 & 69.41 & 75.82 \\ 
				RDC           	   & 81.36 & 78.94 & 77.11 & 80.66 & 82.40 & 78.47 & 75.96 & 79.39 & 75.63 & 66.20 & 69.59 & 70.21 & 76.33 \\
				R-DANN         	   & 81.38 & 79.30 & 77.57 & 80.70 & 82.71 & 78.38 & 76.00 & 79.18 & 76.18 & 66.64 & 69.83 & 69.62 & 76.46 \\
				VRADA                  & 82.12 & 80.68 & 77.71 & 82.24 & 83.09 & 78.82 & 76.27 & 80.00 & 76.28 & 68.20 & 70.01 & 71.34 & 77.23 \\
				\hline
				SASA-$\bm{\alpha}$             & 84.62 & 81.02 & 79.89 & 83.36 & 84.12 & 80.78 & 76.78 & 80.72 & 78.37 & 68.65 & 70.62 & 72.23 & 78.47 \\
				SASA-$\bm{\beta}$             & 83.68 & 81.47 & 78.36 & 82.70 & 84.36 & 81.20 & 77.14 & 80.52 & 77.86 & 68.23 & 70.35 & 72.57 & 78.20 \\
				
				SASA                  & \textbf{85.03}     & \textbf{82.91}     & \textbf{80.32} & \textbf{83.82} & \textbf{85.20} & \textbf{82.03} & \textbf{77.83} & \textbf{81.10} & \textbf{78.93} & \textbf{69.02} & \textbf{70.96} & \textbf{72.76} & \textbf{79.16} \\
				\hline
		\end{tabular}}
		
		\caption{AUC score(\%) on in-hospital mortality prediction.}
		\label{tab:mimic3}
	\end{table*}
	
	
	\subsubsection{Boiler Fault Detection Dataset.}
	The boiler data consists of sensor data from three boilers from 2014/3/24 to 2016/11/30. There are 3 boilers in this dataset and each boiler is considered as one domain. The learning task is to predict the \textit{faulty blowdown valve} of each boiler. Since the fault data is very rare. It's hard to obtain the fault samples in the mechanical system. So it's important to utilize the labeled source data and unlabeled target data to improve the model generalization. 
	
	\subsubsection{Air Quality Forecast Dataset.}
	The air quality forecast dataset\cite{zheng2015forecasting} is collected in the Urban Air project\footnote{https://www.microsoft.com/en-us/research/project/urban-air/} from 2014/05/01 to 2015/04/30, which contains air quality data, meteorological data, and weather forecast data, etc. The dataset covers 4 major Chinese cities: Beijing (B), Tianjin (T), Guangzhou(G), and Shenzhen(S). We employ air quality data as well as meteorological data to predict the PM2.5. We choose the air quality station with the least missing value and take each city as a domain. We use this dataset because the air quality data is common and the sensors in the smart city systems usually contain complex causality. The association among sensors are often sparse, which is suitable for our model. 
	
	\subsubsection{In-hospital Mortality Prediction Dataset.}
	MIMIC-III\cite{johnson2016mimic,che2018recurrent}\footnote{https://mimic.physionet.org/gettingstarted/demo/} is another published dataset with de-identified health-related data associated with more than forty thousand patients who stayed in critical care units of the Beth Israel Deaconess Medical Center between 2001 and 2012. 
	It's the benchmark of time series domain adaptation in VRADA\cite{DBLP:conf/iclr/PurushothamCNL17}.
	Similar to Purushotham et al.\cite{purushotham2018benchmarking}, we choose 12 time series (such as Heart Rate, Temperature, Systolic blood pressure, etc) from 35637 records. In order to prepare the in-hospital mortality prediction dataset for time series domain adaptation, we split the patients into 4 groups according to their age (Group1: 20-45, Group2: 46-65, Group3: 66-85, Group4: \textgreater 85).
	\subsection{Baseline}
	 
	\noindent\textbf{LSTM\_S2T.} LSTM\_S2T uses the source domain data to train a vanilla LSTM model and applies it to the target domain without any adaptation (S2T stands for source to target). It's expected to provide the lower bound performance.\\
	\textbf{R-DANN.} {R-DANN \cite{da2020remaining} is an unsupervised domain adaptation architecture proposed in \cite{ganin2015unsupervised} with GRL (Gradient Reversal Layer) on LSTM, which is a straightforward solution for time series domain adaptation.\\
	\textbf{RDC.} Deep domain confusion is an unsupervised domain adaptation method proposed in \cite{tzeng2014deep} which minimizes the distance between the source and target distributions by employing Maximum Mean Discrepancy (MMD). Similar to the aforementioned R-DANN, we use LSTM as the feature extractor for time series data.\\
	\textbf{VRADA.} VRADA \cite{DBLP:conf/iclr/PurushothamCNL17} is a time series unsupervised domain adaptation method which combines the GRL and VRNN \cite{chung2015recurrent}.
	
	For a fair comparison, the total numbers of parameters of all the baselines and our method are about equal, which is shown in Table \ref{tab:params}. We use the same parameter combination on each dataset and also apply three different random seeds to each experiment. 
	\begin{table}
		\centering
	\begin{tabular}{c|ccc}
		\hline
		\small{Method}  & Boiler & Air & MIMIC-III \\
		\hline
		\small{LSTM\_S2T} 	           & 82924 & 46191 & 72106  \\ 
		\small{RDC}              	   & 82924 & 46191 & 72106 \\
		\small{R-DANN}            	   & 82652 & 46183 & 71479  \\
		\small{VRADA}                  & 83532 & 45898 & 72784  \\
		\hline
		\small{Ours}                   & 82322 & 45636 & 71402\\
		\hline
	\end{tabular}
		\caption{Total numbers of parameters of all the methods in different datasets.}
		\label{tab:params}
	\end{table}
	
	\subsection{Model Variants}
	In order to verify the effectiveness of each component of our model, we further devise the following model variants.
	\begin{itemize}
    	\item SASA-$\bm{\alpha}$: We remove $\mathcal{L}_\alpha$ to verify the usefulness of the segment length restriction loss.
	    \item SASA-$\bm{\beta}$: We remove $\mathcal{L}_\beta$ to verify the usefulness of the sparse associative structure alignment loss.
	\end{itemize}

	\subsection{Result}
	\subsubsection{Results on Boiler Fault Detection.}
	The AUC result in the boiler fault detection dataset is shown in Table \ref{tab:boiler}. Our SASA model significantly outperforms other baselines on all the tasks. It's worth mentioning that our sparse associative structure alignment model promotes the AUC score substantially on harder transfer tasks, e.g. 1 $\rightarrow$ 2 and  3 $\rightarrow$ 2, which are respectively improved by 3.95 and 2.56 points compared with VRADA. On some easy tasks such as  1 $\rightarrow$ 3, 2 $\rightarrow$ 3, and 3 $\rightarrow$ 1, though the other baselines perform well, our method still achieves a comparable result. We also conduct the Wilcoxon signed-rank test \cite{wilcoxon1945individual} on the reported score, the p-value is 0.027, which means that our method significantly outperforms the baselines with the p-value threshold 0.05.
	
	\begin{table}[H]
		\centering
		\begin{tabular}{p{1.4cm}|p{0.55cm}p{0.55cm}p{0.55cm}p{0.55cm}p{0.55cm}p{0.55cm}p{0.55cm}p{0.55cm}}
			\hline
			\small{Method}&\small{1$\rightarrow$2}&\small{1$\rightarrow$3}&\small{3$\rightarrow$1}&\small{3$\rightarrow$2}&\small{2$\rightarrow$1}&\small{2$\rightarrow$3}&\small{Avg} \\
			\hline
			\small{LSTM\_S2T} 	           & \small{67.09} & \small{94.54} & \small{93.14} & \small{56.09} & \small{84.99} & \small{91.31} & \small{81.19} \\ 
			\small{RDC}              	   & \small{68.29} & \small{94.65} & \small{93.38} & \small{57.32} & \small{85.31} & \small{92.57} & \small{81.92} \\
			\small{R-DANN}            	   & \small{67.71} & \small{94.69} & \small{93.92} & \small{58.53} & \small{85.67} & \small{91.66} & \small{82.03} \\
			\small{VRADA}                  & \small{67.59} & \small{94.88} & \small{93.65} & \small{60.59} & \small{85.96} & \small{92.62} & \small{82.55} \\
			\hline
			\small{SASA-$\bm{\alpha}$ }                  & \small{70.83} & \small{95.86} & \small{94.63} & \small{60.76} & \small{87.27} & \small{93.28} & \small{83.77} \\
			\small{SASA-$\bm{\beta}$ }                  & 
			\small{69.76} & \small{95.01} & \small{94.56} & \small{61.31} & \small{86.78} & \small{92.84} & \small{83.38}\\
			
			\small{SASA}                  & \small{\textbf{71.54}} & \small{\textbf{96.39}} & \small{\textbf{94.77}} & \small{\textbf{63.15}} & \textbf{\small{87.76}} & \small{\textbf{93.59}} & \small{\textbf{84.53}} \\
			\hline
		\end{tabular}
		
		\caption{AUC score(\%) on boiler fault detection.}
		\label{tab:boiler}
	\end{table}
	
	\subsubsection{Results on Air Quality Forecast.}
	Similar to the result in the boiler fault detection dataset, our model also achieves great results and outperforms all the other baselines on all tasks, which is reported in Table \ref{tab:air_quality}. According to the result, we can observe that: 1) The performance between the closer cities is better than that of farther cities. For example, since the distance between Beijing and Tianjin is smaller than the distance between Beijing and Guangzhou and between Beijing to Shenzhen, the promotion of $B\rightarrow T$ is better than that of $B\rightarrow G$ and $B\rightarrow S$. This is because the cities pair with closer distance share more common associative structure. 2) Our method still achieves the best result even the source city is far away from the target city, e.g. Beijing and Shenzhen, this phenomenon reflects that our sparse associative structure alignment model well extracts the associative structure among different variables. 3)The performance is not so notable compared with other tasks when Shenzhen is taken as the target domain. This is because the label value of this domain is much lower than other domains. We also conduct the Wilcoxon signed-rank test \cite{wilcoxon1945individual} on the reported score, the p-value is 0.002, which means that our method significantly outperforms the baselines with the p-value threshold 0.05.
	
	\subsubsection{Results on In-hospital Mortality Prediction.}
	We also testify our model on MIMIC-III dataset, which is chosen as the benchmark of time series domain adaptation in \cite{DBLP:conf/iclr/PurushothamCNL17}. We choose 12 variables described in \cite{purushotham2018benchmarking} and reproduce a similar result of VRADA. As shown in Table \ref{tab:mimic3}, our model overpasses the other comparison methods on all the transfer tasks. Some domain adaptation tasks such as 2 $\rightarrow$ 1 and 3 $\rightarrow$ 2 are even improved by 2.91 and 2.11 points respectively. Furthermore, we also find that the performance between similar domains like 1 and 2, 2 and 3, as well as 3 and 4 are better than others. 
	We also conduct the Wilcoxon signed-rank test \cite{wilcoxon1945individual} on the reported score, the p-value is 0.0022, with the p-value threshold 0.05.
	
	\subsection{Ablation Study and Visualization}
	\subsubsection{The study of the usefulness of the segment length restriction.} In order to verify the effectiveness of sparse associative structure alignment, we remove $\mathcal{L}_\alpha$ and the model is named \textbf{SASA}-$\bm{\alpha}$. Compared the result of \textbf{SASA} and \textbf{SASA}-$\bm{\alpha}$, we can find that the performance of \textbf{SASA}-$\bm{\alpha}$ drops. This is because of $\alpha$ represents the probability of the length of a segment. And the duration of segments varies with domains. With the restriction of $\bm{\alpha}$, we can exclude the influence of domain-specific segments duration.
	
	\subsubsection{The study of the effectiveness of sparse associative structure alignment.} 
	In order to verify the effectiveness of the segment length restriction, we remove the sparse associative structure alignment loss. According to the experiment result of \textbf{SASA}-$\bm{\beta}$, we can find that the performance of \textbf{SASA}-$\bm{\beta}$ is worse than standard \textbf{SASA}.  This is because the sparse associative structure has been extracted, which is also more robust than that of normal feature extractor. But the reserved domain-specific associative relationships lead to the suboptimal result. 
	Note that \textbf{SASA}-$\bm{\beta}$ is still better than the other baselines. 
	This is because  $\mathcal{L}_\alpha$ aligns the offsets between different domains, which benefits to extracting sparse associative structure for adaptation.
	

	\subsubsection{Visualization of Aligned Structure.}
	\begin{figure}[t] 
		\centering
		\subfigure{
			\includegraphics[width=0.9\columnwidth,height=0.35\columnwidth]{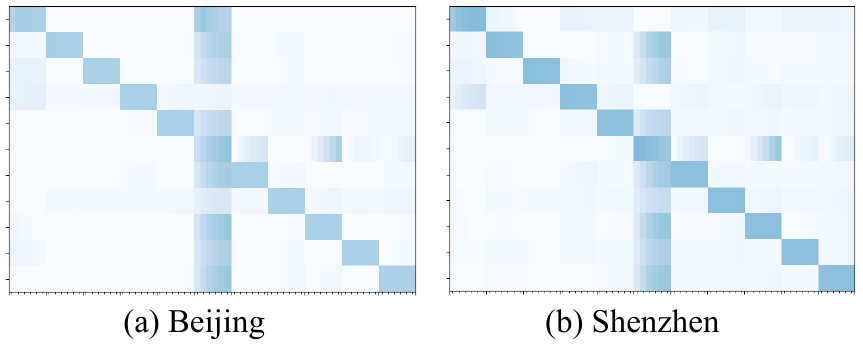}
			\label{fig:transfer}
		}
			\caption{The illustration of visualization of correlation structure adjacent matrix of Beijing $\rightarrow$ Shenzhen. Deeper the color is, the stronger the relationship is. We can find that the structure is sparse.}
		\label{fig:causal_structure}
	\end{figure}

	To further investigate our approach, we perform the visualization of aligned sparse associative structure over the air quality dataset and attempt to extract the common sparse associative structure, which is shown in Figure \ref{fig:causal_structure}. The visualization shows the sparse associative structures of Beijing and Shenzhen respectively. The deeper the color is, the stronger the association between two variables.
	we can find that (1) the associative structures from different domains are very sparse. (2) the sparse associative structures from the source and the target domains have many shared associative relationships, which means that similar sparse associative structure is shared in different domains. 

	\section{Conclusion}
    This paper presents a sparse associative structure alignment model for time series unsupervised domain adaptation. In our proposal, a sparse associative structure discovery method, equipped with an adaptive summarization of the series segments,  is used to extract the structure of the time series, and an MMD based structure alignment method is used to transfer the knowledge from the source domain to the target domain. The success of the proposed approach not only provides an effective solution for the time-series domain adaptation task, but also provides some insightful results on what is transferable on the time-series data.

    \section{Acknowledgments}
    This research was supported in part by Natural Science Foundation of China (61876043, 61976052), Science and Technology Planning Project of Guangzhou (201902010058).
    

\section{Statement About The Potential Ethical Impact}
The Time series domain adaptation model, which is combined with the correlative relationship, is more robust than the existing methods and yields significant performance in unlabeled test data, which can apply in mechanical systems, smart cities as well as healthcare. The positive implications of applying our method include:
\\
(1) Significant improvement of unsupervised domain adaptation for time series data, which reduces the requirement for manually acquiring labeled data and make the machine learning model available for use the in low-resource settings. \\
(2) Unsupervised domain adaptation for time series data is beneficial to mitigate overfitting.
\\
However, the negative implications of the increasingly powerful artificial intelligence technology should not be ignored. These technologies lack interpretability so it's hard to be trusted sometimes. Our method can figure out this circumstance to some extent, but it can be better if we take causality into account.

\bibliography{aaai}

\end{document}